\documentclass[a4paper]{article}

\usepackage[english]{babel}
\usepackage[utf8]{inputenc}
\usepackage{amsmath,amsthm,amsfonts, authblk}
\usepackage{graphicx}
\usepackage[colorinlistoftodos]{todonotes}
\usepackage{tikz-cd}
\usepackage{pgf,tikz}
\usepackage{mathrsfs}
\usetikzlibrary{arrows}

\newcommand{\R}{\mathbb{R}}

\newcommand{\Ell}{\mathcal{L}}
\newcommand{\Ess}{\mathcal{S}}
\newtheorem*{que}{Question}

\title{A Topological Kinematic Workspace Analysis of the Canfield Joint}

\author[1]{Robert Short}
\author[2]{Alan Hylton}
\affil[1]{Lehigh University, Bethlehem, PA}
\affil[2]{NASA Glenn Research Center, Cleveland, OH}
\date{\today}

\begin{document}
\maketitle

\begin{abstract}
We use topological techniques to do a workspace analysis of the Canfield Joint, a mechanical linkage constructed with two plates connected by three legs.  The Canfield Joint has three degrees of freedom and can be controlled using three actuators attached to the base in strategic positions.  In the process of performing the workspace analysis, we describe a new method of controlling the Joint which includes elements of both forward and inverse kinematics.  This control process is then used to answer the question of how the workspace of the joint changes in the possibility of a failure mode where one degree of freedom is lost.

%Detail on Canfield Joint (sent 1), 3 legs, 3 dof
%(sent 1.5) Control of CJ
%(last sentence) Failure mode where you lose a dof
%Frames from Andy's movie of Canfield Joint
\end{abstract}

\section{Introduction}

The Canfield Joint is a linkage initially constructed by Stephen Canfield in his PhD thesis \cite{Canfield}.  He designed the joint in order to emulate the motion of a typical human wrist.  However, its properties have proven useful in many different applications.  One of the most useful properties of the joint is the freedom of its workspace.  The joint can be pointed at every point located on a hemisphere, meaning the effective workspace is hemispherical.  This first attracted NASA's attention as it is convenient for pointing rocket engines precisely.  However, that was not the last time NASA would find an application of the joint.

\begin{figure}
\centering
\includegraphics[width = 0.7\textwidth]{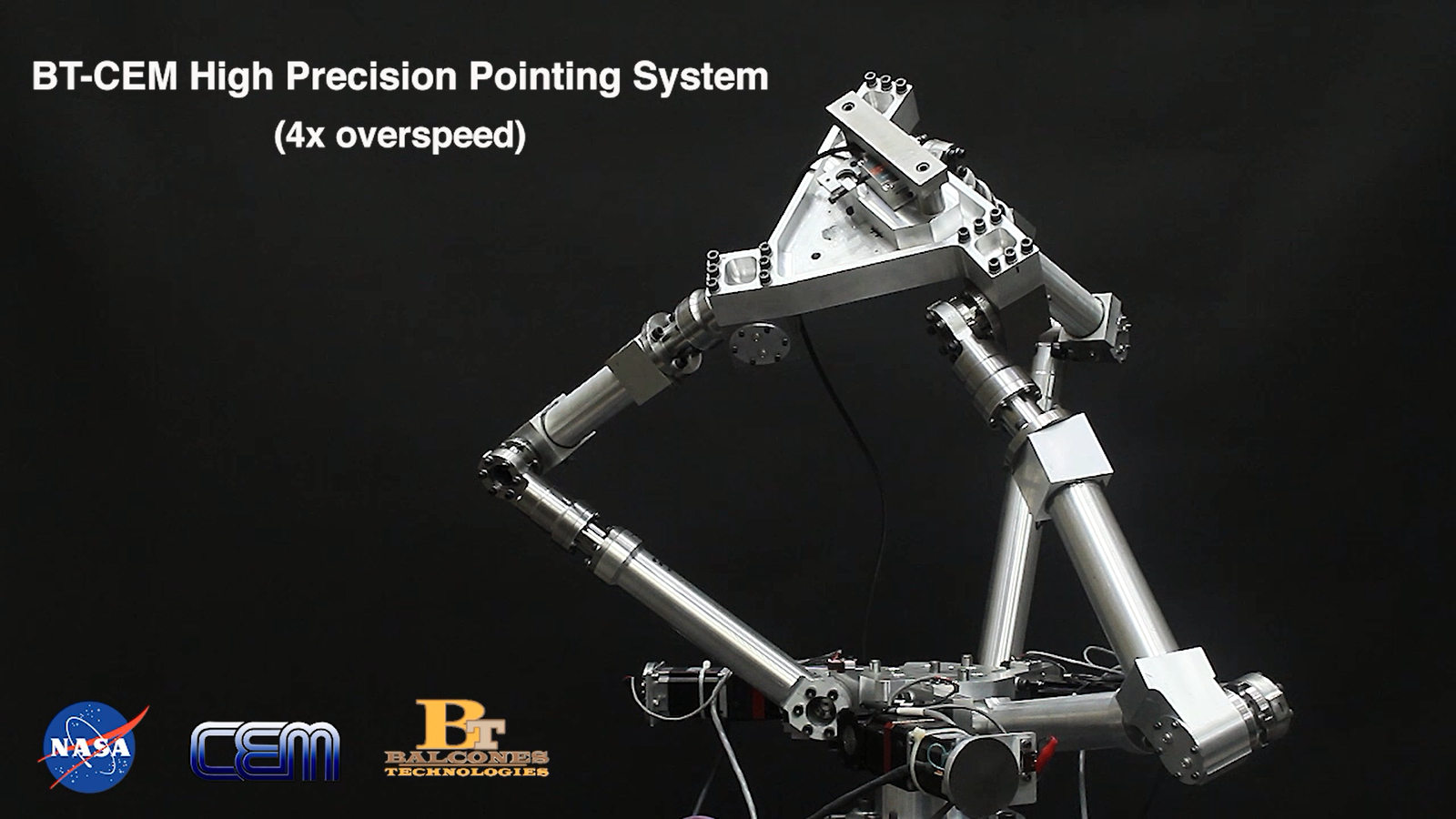}
\caption{\label{iROC}This image of iROC's Canfield Joint was provided by Andrew Rocha.}
\end{figure}

Researchers on NASA's Integrated Radio and Optical Communications (iROC) project are currently developing technologies for deep space radio and optical communications.  The goal of the iROC project is to use radio and optical communication to transmit data between satellites around Mars and the Earth.  In the iROC project, the Canfield Joint has found a new purpose due to its unique design features.  The iROC project intends to send lasers from Mars to Earth, a distance between 0.4 and 2 AU depending on the relative locations of the planets.  The great distance causes the laser's footprint on Earth to be approximately the size of Texas.  Thus, a joint with a wide workspace that can be moved precisely was needed.  The Canfield Joint plays two important roles as the pointer of the antenna assemblies for the satellite.  First, the hemispherical workspace is necessary, along with the ability to point precisely in whatever direction is needed.  Second, the construction of the joint leaves space for simple, flexible wires to connect the satellite to the antenna unhindered by mechanical components, allowing for a much lower wiring weight.  A picture of the Canfield Joint being used by the iROC team is provided in Figure \ref{iROC}.  While these are two very useful properties, there are other claimed properties that could be critically important, but have yet to be tested.
%Picture of CJ w/ cables if it exists

One such property that has been claimed of the Canfield Joint is the workspace's robustness if one of the arms breaks.  When we say an arm ``breaks'' we do not mean that there is a disconnect in the joint itself.  The Canfield Joint is traditionally controlled by motors located at the base angles of the joint.  We say that an arm ``breaks'' when one of these motors fails and we are unable to change the ``broken'' base angle.  We can simulate this in the workspace by fixing one of the base angles and examining how the workspace is restricted.  The motivation of this paper is to be able to give a more precise description of this new workspace.  In other words, we seek to answer the following question:

\begin{que}
What is the kinematic workspace of the Canfield Joint if one of its arms breaks?
\end{que}

In the process of answering this question, we will proceed in the following way.  First, we will introduce more precisely the construction of the Canfield Joint, as well as some of the necessary mathematical background.  We will be using techniques from topology discussed in Kevin Walker's undergraduate thesis \cite[\S 2]{Walker}, and so it will prove useful to introduce those techniques.  We will turn our attention to a single arm of the Canfield Joint which we can think of as the arm that may ``break''.  In the process of performing our analysis, we discover an alternate control scheme for the Canfield Joint which then permits us some simulation-based answers to the question.%With any luck, we will have experimental verification of some of the results.

\section{The Canfield Joint}

We will begin by detailing the necessary components of a construction of the Canfield Joint.  For readers seeking further details, we are essentially summarizing \cite[Ch. 3]{Canfield}.

\begin{figure}
\centering
\includegraphics[width = 0.6\textwidth]{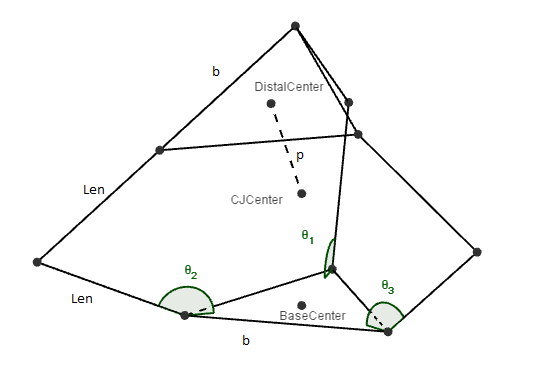}
\caption{\label{Construction}Key measurements of the Canfield Joint.  }
\end{figure}

\subsection{Construction}

The Canfield Joint as constructed in Dr. Canfield's thesis consists of two triangular plates connected by three arms.  Each arm is then divided into two pieces of equal length.  While in reality the midpoints of the arms are connected by a hinge joint and two free rotation joints perpendicular to each piece of the arm, Canfield emulates this with a spherical joint.  As this is equivalent, it is the approach we will take here.  The arms connect to the plates with hinge joints whose axes are tangent to the circumscribed circle of the base triagle.  This joint is controlled by specifying the angles that the arms make with the base plate.

We say that $\ell$ (denoted Len in Figure \ref{Construction}) is the length of each piece of the arm, saying each arm is two bars of length $\ell$ connected by a spherical joint.  The straight-line distance between any two hinges on the same plate is $b$, and it is fixed for any pair of such hinges.  We call the plate which is fixed the ``base plate'' and the plate that moves around the ``distal plate''.  The three angles between each arm and the base plate are given by $\theta_1$, $\theta_2$ and $\theta_3$.  We will use the subscripts 1, 2, and 3 to distinguish between the three arms agreeing with the corresponding angle numbers.  A picture of this construction if provided in Figure \ref{Construction}.

One other vital aspect of the construction is what is known as the plunge distance.  To find this, we first need what we refer to as the Center of the joint.  First, compute the centers of each plate.  Then, construct normal lines to each plate.  The Center of the joint is the intersection of these two lines, which exists by the construction of the joint.  We call the plunge distance, $p$, the distance along the normal line from the Center of the joint to the center of either plate.  By the symmetry inherent in the construction, this distance is the same whether going from the Center of the joint to the base or distal plates.

\begin{figure}
\centering
\includegraphics[width = 0.6\textwidth]{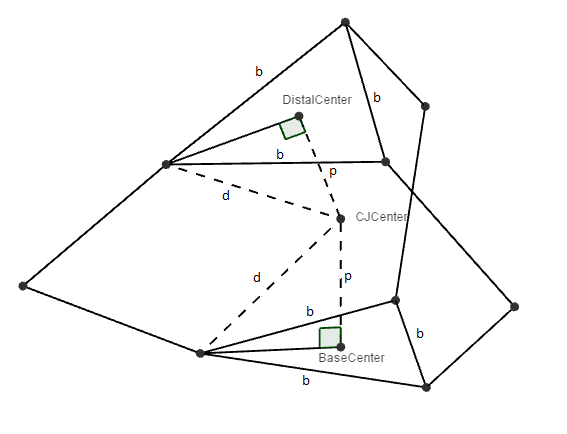}
\caption{\label{dDefn}How the parameter $d$ is defined.   }
\end{figure}

There is one more distance to be measured which has not been referred to in previous constructions.  This is the distance from the Center of the joint to each hinge.  Again, this is fixed by the symmetry of the construction.  We label this $d$, but it is given by the convenient formula (thanks to Pythagoras) $d = \sqrt{p^2+b^2/3}$.  The right triangle this is derived from is constructed in Figure \ref{dDefn}.  This measurement will prove useful when we begin topological considerations.

Note that for any construction of a Canfield joint, the lengths $\ell$ and $b$ are fixed.  Since any Canfield Joint is controlled by $\theta_1$, $\theta_2$, and $\theta_3$, it is possible to write all other variables in terms of these five measurements.  Some of the details on this can be found in \cite[Ch. 4]{Canfield}.

%--Reference Canfield's Thesis for Joint Construction \\

%--Use Pythagorean Theorem to construct $d$ and append bounds to it. \\

\subsection{Some Physical Restrictions}
Due to how the joint is constructed, and the fact that it is an inherently physical object, we can place some bounds on the previously defined measurements.  These bounds will be useful later on in restricting our attention to valid configurations of the joint.

For $p$, we first note that we do not want the distal plate to pass through the base plate.  As such, we will make certain that $p \geq 0$. Moreover, with some applications of the triangle inequality, we get that $p \leq \ell$.  In essence, this is due to there being at least one arm that cannot reach the distal plate when $p > \ell$.  We can then translate these bounds to bounds on $d$.  When done, we get that 
\begin{equation}
\frac{b}{\sqrt{3}} \leq d \leq \sqrt{\ell^2+\frac{b^2}{3}}.
\end{equation} 

The other restriction to consider restricts each of the base angles.  We define $\theta_i$ to be the angle at the base hinge between the midpoint of arm $i$ and the center of the base plate, as depicted in Figure \ref{Construction}.  As such, we have that $0 \leq \theta_i < 2\pi$ for each $i=1,2,3$.  Our typical starting position of the joint occurs when $\theta_i = \pi$ for each $i=1,2,3$.  We can impose stricter restrictions on each $\theta_i$ which will become apparent as we unravel the workspace analysis.

These restrictions are important so that our final model accurately reflects reality.  However, when we do the topological workspace analysis, we will initially not consider physical restrictions.

%--Use physicality of the joint to describe certain bounds

\section{Mathematical Background}
In taking a topological view of robotics, mathematicians have combined and transformed ideas from both topology and robotics to piece together a coherent theory.  In this section, we introduce many of the phrases and ideas relevant to this work.

\subsection{The Configuration Space to Workspace Map}
Let $\Ell$ denote a mechanical linkage.  The configuration space of $\Ell$, often denoted $C(\Ell)$, is the set of configurations of the linkage.  It is given a topology relevant to the construction of the linkage, and this turns the set $C(\Ell)$ into a topological space.  In many cases, this can be made into a metric space.

For any linkage, $\Ell$, there is some Euclidean space $\R^n$ in which we can embed $\Ell$.  If we choose a fixed $\R^n$ and a method of embedding a configuration in $C(\Ell)$ into $\R^n$, then for a point $x \in \Ell$, one can track the possible locations in $\R^n$ that $x$ can occupy.  This set of possible locations is called the workspace of $x$, and it is often denoted $W_{\Ell}(x)$.  If either the point $x$ or linkage $\Ell$ is understood in context, we will suppress the notation in the respective way ($W_\Ell$, $W(x)$, or even just $W$).  Fortunately, this set inherits its topology as a subspace of $\R^n$, so there is always a metric structure on the workspace.

There is a natural function $f_x: C(\Ell) \to W_{\Ell}(x)$ sending each configuration of $\Ell$ to the location of $x$ in that configuration.  This function is what we will exploit through the remainder of the paper in order to draw out the desired workspace.

%--Config $\to$ Workspace Map \\

\subsection{Factoring Through Sublinkages}
A sublinkage $\Ess$ is a linkage constructed from some subset of the pieces in the original linkage.  While it is not necessarily the case that $C(\Ess) \subseteq C(\Ell)$, there is a function $\eta_\Ess : C(\Ell) \to C(\Ess)$ given by sending the configuration of the pieces of $\Ess$ in the full linkage to the same configuration in $C(\Ess)$.

Notice that if $x \in \Ess$, we get a second map $g_{x,\Ess}:C(\Ess) \to W_{\Ess}(x)$.  Unlike what happens with configuration spaces, we have that $W_{\Ell}(x) \subseteq W_{\Ess}(x)$.  So, when we examine the inverse image of $W_\Ess(x)$ under $g_{x,\Ess}$, denoted $g_{x,\Ess}^{-1}(W_\Ess(x))$, we can restrict our attention to $g^{-1}_{x,\Ess}(W_{\Ell}(x))$.  This, it turns out, is a very important construction.

The reason this construction is valuable is that we can construct a commutative diagram connecting all of these pieces together.  Let $\Ell$ be a linkage with sublinkage $\Ess$ where $x \in \Ess$.  Then, the following diagram commutes:

\begin{center}
\begin{tikzcd}
C(\Ell)\arrow{dd}{f_x} \arrow{rd}{\eta_\Ess }\\
&C(\Ess) \arrow {ld}{g_{x,\Ess}}\\
W_{\Ess}(x)
\end{tikzcd}
\end{center}

While we initially defined $f_x:C(\Ell) \to W_{\Ell}(x)$, it is easy to see that $f_x(C(\Ell))\subseteq W_\Ess(x) \subseteq W_\Ell(x)$, i.e. the image of $f_x$ is contained in $W_\Ess(x)$ which we know is itself contained in $W_\Ell(x)$.  In the sequel, we will need to consider multiple sublinkages simultaneously.  We will see why shortly.

%--Factoring through other config spaces \\

\subsection{Going Backwards}
One way to understand $C(\Ell)$ is to examine carefully the inverse image of the workspace function $f_x:C(\Ell) \to W_{\Ell}(x)$ at each point in the workspace.  Doing this, however, can be tricky; it may not be easy to describe the configuration space directly like this.  Topologists have been able to circumvent, and thus solve, this issue by reducing the problem to simpler cases.  In particular, we consider sublinkages and describe $f_x^{-1}(W_{\Ell}(x))$ using the factorization property described above.

The process uses the following methodology.  First, we break our linkage $\Ell$ into sublinkages $\Ess_i$ for $i=1, \dots ,n$ such that $\bigcup_{i=1}^n \Ess_i = \Ell$, and the pairwise intersections coincide with the total intersection which is $\bigcap_{i=1}^n \Ess_i = \Ess^* \cup \{x\}$ for some sublinkage of fixed links $\Ess^*$.  Then, we get that
\begin{eqnarray*}
C(\Ell) & =f_x^{-1}(W_{\Ell}(x)) & = f_x^{-1}\bigg(\bigcap_{i=1}^n W_{\Ess_i}(x)\bigg),\\
 & & = \bigcap_{i=1}^n\big[ f_x^{-1}( W_{S_i}(x) ) \big], \\
 & & = \bigcap_{i=1}^n\big[ \eta_{\Ess_i}^{-1}( g_{x,\Ess_i}^{-1}( W_{\Ess_i}(x) ) ) \big].\\
\end{eqnarray*}

While this may appear to be a very complicated expression, the essence of it is easily stated.  We break our linkage down into sublinkages in the hopes that workspace analysis will be easier there.  Since, in our original configuration space, the map must factor through each of the sublinkages, we can say that it must satisfy all of the conditions imposed by said sublinkages.  That is, the workspace of $\Ell$ is the intersection of the $\Ess_i$ workspaces.  Then, once we have done this for a comprehensive collection of sublinkages, we can combine these results to determine the configuration space of the full linkage.  This is a way to work up the diagram in section 3.2 and recover information about the configuration space.

%--Show that intersection is Correct thing to Examine \\

\subsection{Planar 4-bar Linkage:  An Example}

\begin{figure}
\centering
\includegraphics[width = 0.4\textwidth]{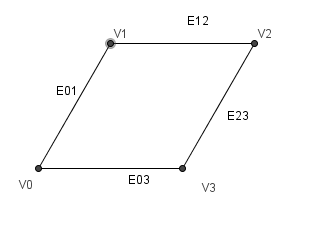}
\caption{\label{Planar}An image of the Planar 4-bar Linkage.   }
\end{figure}

As an example, consider a 4-bar cyclic linkage $\Ell$ embedded in the plane ($\R^2$) as depicted in Figure \ref{Planar}.  We assume that one of the joints is fixed at the origin, and that all of the arms have the same length with one arm fixed along the $x$-axis.  Ordering the joints $V0,V1,V2,V3$ with $V0$ fixed at the origin and choosing a direction to orient the edges gives us a way to talk about the workspace of various points.  Also, note that the joints must all be hinge joints due to $\Ell$ being embedded in a plane.  Using this, we assume that the edge $E03$ is fixed to the $x$-axis.  To simplify the analysis, we will assume that the joints and arms are permitted to move through each other, so we do not need to worry about preventing collisions.

We will consider the workspace of the point $V2$, $W_{\Ell}(V2)$.  To do this, we can divide $\Ell$ into two sublinkages; $\Ess_1$ will consist of the joints $V0,V1,V2$ and their connecting arms while $\Ess_2$ will consist of the joints $V0,V3,V2$ and their connecting arms.  Note that $\Ess_1 \cup \Ess_2 = \Ell$ and $\Ess_1 \cap \Ess_2 = \{V0\} \cup \{V2\}$.  In the prior notation, $\{V2\} = \{x\}$, the point defining our workspace, and $\{V0\} = \Ess^\ast$, the fixed sublinkage contained in each of the sublinkages considered.

\begin{figure}
\centering
\includegraphics[width = 0.3\textwidth]{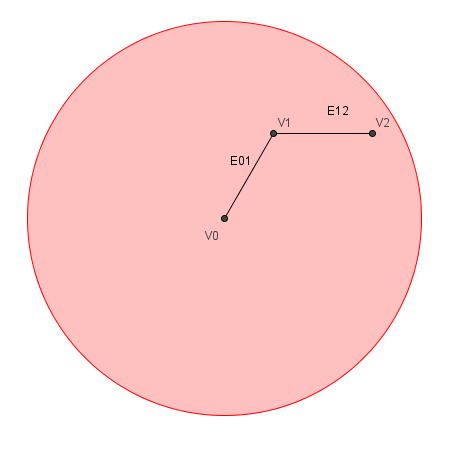}
\caption{\label{Workspace1}An image of the Workspace $W_{\Ess_1}(V2)$   }
\end{figure}

When we go to consider $C(\Ess_1)$, we can think of the two arms as being vectors in $\R^2$ of unit length.  As such, we have a circle's worth of options for the location of each arm, and then we add the two vectors together to get each configuration.  This means that $C(\Ess_1)$ is homeomorphic to $S^1\times S^1$, the standard torus!  However, when we consider the map to $W_{\Ess_1}(V2)$, we notice that the workspace is swept out by rotating a circle of radius 1 centered at $(1,0)$ about the origin.  A picture of the resulting space is given in Figure \ref{Workspace1}.

Considering the inverse image of each point in the workspace $W_{\Ess_1}(V2)$ under $f_{V2}$, we see that, apart from the origin, there are only finitely many configurations that correspond to any point in the workspace.  For example, if $b$ is a point on the boundary circle, then $f_{V2}^{-1}(b)$ is a single configuration, in particular $f_{V2}^{-1}(b)$ would be the straight line configuration in which both vectors are pointed toward $b$.  The only remaining case is when $p$ is a point on the interior of the circle, but not the origin.  Then, we can see that the number of configurations in $f_{V2}^{-1}(p)$ is 2.  Thinking of the workspace as the region swept out by rotating circles, we can easily find two distinct circles that intersect a fixed $p$, and then an interested reader could demonstrate why there cannot be any additional circles that intersect $p$.

The origin, however, is a special case.  It is helpful to first notice that when one vector is the negative of the other, we have that $V2$ is back at the origin.  Since this occurs precisely once for each choice of the first vector in $C(\Ess_1)$, we see that this occurs for a whole circle's worth of points, and thus $f_{V2}^{-1}((0,0))$ is homeomorphic to a circle, $S^1$. 

\begin{figure}
\centering
\includegraphics[width = 0.3\textwidth]{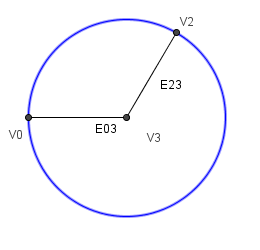}
\caption{\label{Workspace2}An image of the Workspace $W_{\Ess_2}(V2)$   }
\end{figure}

On the other hand, when we look at $C(\Ess_2)$, we can again think of the two arms as vectors in $\R^2$ of unit length.  However, in this case, one of the vectors is fixed to fall along the $x$-axis, so the configuration space is determined by a single vector in $\R^2$ of unit length.  That is, $C(\Ess_2)$ is homeomorphic to $S^1$, which is just a circle!  Moreover, when we examine $W_{\Ess_2}(V2)$, we have exactly the circle displayed.  A picture of $W_{\Ess_2}(V2)$ is shown in Figure $\ref{Workspace2}$.  Thus, the inverse function analysis is simple here, $f_{V2}:C(\Ess_2) \to W_{\Ess_2}(V2)$ is a one-to-one correspondence.  That is, each location in this workspace corresponds to a single configuration of the sublinkage.

\begin{figure}
\centering
\includegraphics[width = 0.3\textwidth]{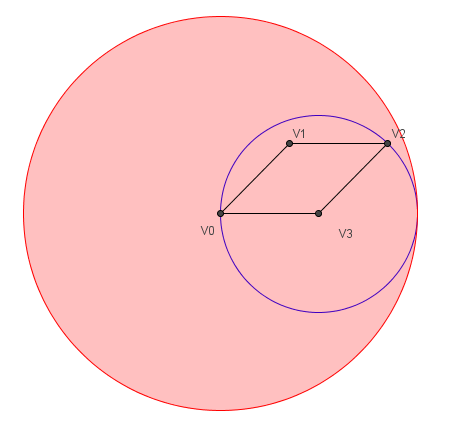}
\caption{\label{PIntersect}Overlaying the workspaces $W_{\Ess_1}(V2)$ and $W_{\Ess_2}(V2)$ makes the intersection clear.   }
\end{figure}

Our final step is to intersect the workspaces and consider the intersections of the inverse images of the corresponding maps.  A picture of the intersection of the workspaces is given in Figure \ref{PIntersect}.  Since $W_{\Ess_2}(V2)$ is contained in $W_{\Ess_1}(V2)$, their intersection is just $W_{\Ess_2}(V2)$.  Thus, $W_{\Ess_2}(V2) = W_{\Ell}(V2)$, and this makes our considerations much simpler.

All three types of points that we considered in the inverse image of $W_{\Ess_1}(V2)$  occur in $W_{\Ell}(V2)$.  Since the inverse images of the points in $W_{\Ess_2}(V2)$ are unique, the only difference in configurations will occur from inverse images of $W_{\Ess_1}(V2)$.  First, we notice that the origin is contained in $W_{\Ess_2}(V2)$.  There is a full circle's worth of configurations in $C(\Ess_1)$ which meet at the origin, and so there is a similar circle in $C(\Ell)$ which maps to the origin in $W_{\Ell}(V2)$.  Similarly, there is exactly one boundary point of $W_{\Ess_1}(V2)$ contained in $W_{\Ell}(V2)$.  There is a unique configuration in $C(\Ess_1)$ which meets at the boundary point, implying that there is a unique configuration in $C(\Ell)$ which sends $V2$ to that boundary point.  Finally, the remaining points in $W_{\Ell}(V2)$ lie neither on the boundary nor at the origin.  Thus, there are two configurations in $C(\Ess_1)$ corresponding to each of these points, implying that there are exactly two configurations in $C(\Ell)$ corresponding to each of these points determined by the configurations in $C(\Ess_1)$.

\begin{figure}
\centering
\begin{tikzpicture}[line cap=round,line join=round,>=triangle 45,x=1.0cm,y=1.0cm]
\draw(8.14*0.5,1.72*0.5) circle (3.84541285169746*0.5cm);
\draw (0.5*11.78,0.5*0.48)-- (0.5*7.273851312311967,0.5*5.466596649069165);
\draw (7.273851312311967*0.5,5.466596649069165*0.5)-- (5.210520515342593*0.5,-0.7710539835522165*0.5);
\draw (5.210520515342593*0.5,-0.7710539835522165*0.5)-- (11.78*0.5,0.48*0.5);
\end{tikzpicture}

\caption{\label{PlanarEx}A space homeomorphic to the configuration space for the planar 4-bar linkage.   }
\end{figure}
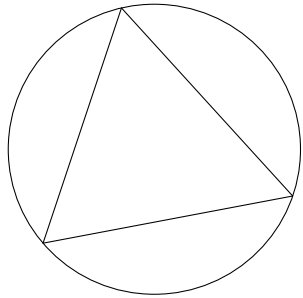

We can carefully piece this information together to determine the configuration space $C(\Ell)$.  Michael Farber gives the description explicitly in \cite[Ch. 1.3, Case H]{Farber}, and we construct a space homeomorphic to $C(\Ell)$ depicted in Figure \ref{PlanarEx}.
%Need to explain this aspect of the construction.  At the very least give some hints for people trying to do the construction (Is doable).

%--Give Ex. from Farber/Walker of 4-bar linkage.

\section{Workspace Analysis}
A previous numerical workspace analysis has been partially done in \cite[Ch. 3.5]{Canfield}.  Since we are using topological methods, we achieve an analysis where we can choose which physical restrictions to implement.
%--Refer to Canfield's Thesis for preliminary numerical analysis. \\

\subsection{Single Arm Analysis}

\begin{figure}
\centering
\includegraphics[width = 0.4\textwidth]{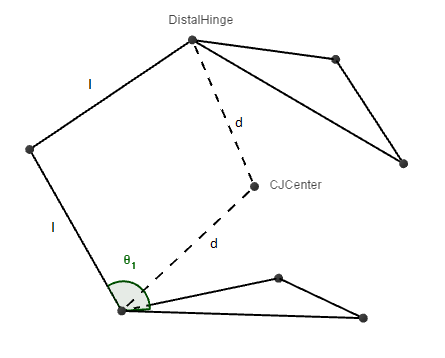}
\caption{\label{Single} The Canfield Joint restricted to a single arm.   }
\end{figure}
%-DMD-Explain why this restriction is useful for answering the question asked.
Since we are trying to answer the question ``What happens to the workspace if one arm breaks?'', we begin by restricting our attention to the configuration space of a single arm.  Recall that we are thinking of this as the arm that may ``break'', so we will care what base angle this arm attains.  Our goal will be to reconstruct the full Canfield Joint workspace using this restriction.  An image of the result of this restriction appears in Figure \ref{Single}.

When we examine the sublinkage consisting of one set of arms, the base plate, and the distal plate, we can recognize a 4-bar linkage hiding in its configuration space.  The four bars are the two arms of length $\ell$ and the two length $d$ segments connecting the base hinge to the center of the joint, and the distal hinge to the center of the joint.  In this 4-bar linkage, we have two true hinges, the distal and base hinges, but the other joints are both ball joints.  The center of the joint is a ball joint since there is no physical restriction to its movement.

Using a similar analysis to the previous example, we can consider the workspace of the distal hinge.  Then, we subdivide the 4-bar linkage $\Ell$ into two sublinkages: the linkage formed by the arms of length $\ell$ which we call $\Ess_{\ell}$ and the linkage formed by the arms of length $d$ which we call $\Ess_d$.  In what follows, we will assume that the workspaces are all the workspaces of the distal hinge in the respective linkages.

\subsubsection{Configuration Space and Workspace of $\Ess_\ell$}
We will first consider $C(\Ess_\ell)$.  Since we are only considering the two length $\ell$ arms, we can break this down into two parameters.  The first is a vector on a circle determined by the base hinge.  The second is a vector on a sphere determined by the possible locations from the ball joint.  Thus, we can see that $C(\Ess_\ell)$ is homeomorphic to $S^1 \times S^2$.  While this is a rather simple description, the workspace is more challenging to describe.

\begin{figure}
\centering
\includegraphics[width = 0.4\textwidth]{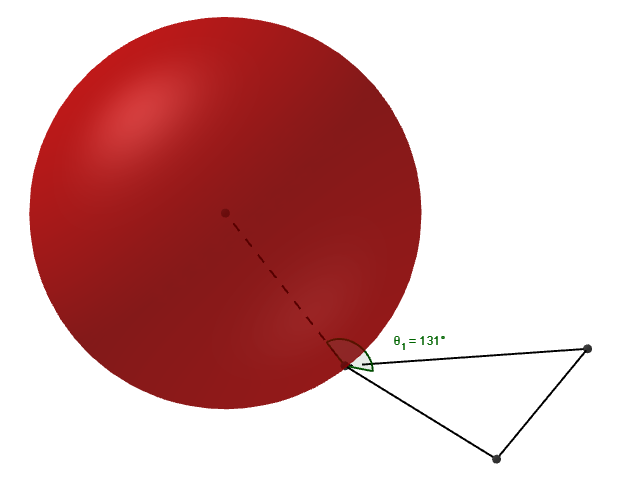}
\caption{\label{Workl}A slice of $W_{\Ess_\ell}$.   }
\end{figure}

We begin by locating the ball joint at the midpoint of the arm.  If we let the origin be the base hinge, then the ball joint can be at any point along a circle of radius $\ell$ about the origin, as depicted in Figure \ref{Workl}.  Without loss of generality, suppose that circle is in the $xy$-plane.  Then, for each point on the circle, there is a sphere's worth of locations for the distal hinge where the sphere is of radius $\ell$.  So, the workspace is the union of all of the spheres generated in this way.  Note that this is a solid object of rotation that one can generate by revolving a circle of radius $\ell$ centered at $(\ell,0,0)$ and passing through the origin about the $z$-axis.

By examining the cross-section given by the solid circle, we can recover the inverse image of each point in the workspace as it is related to the configuration space.  Notice that, since the construction of the workspace is by revolving a sphere about the axis, each point in the interior of the circle has two spheres that map to it, meaning that there are two configurations that correspond to each interior point.  Also, for most of the points on the perimeter of the circle (other than the origin), there is a unique configuration associated with those workspace points as there is only one sphere that contains those boundary elements.  The most curious case occurs at the origin, where there must occur a circle's worth of points in the inverse image.  This is because the origin is reached by all configurations where the second vector is precisely the negative of the first vector.

\subsubsection{Configuration Space and Workspace of $\Ess_d$}
Since we are only considering the $d$-length arms, the configuration space $C(\Ess_d)$ is defined by two parameters.  The first is naturally $d$, since this changes with the plunge distance.  In fact, we already have the bounds on $d$ given by Equation 1.  The other parameter is the location of the second arm, since the plunge distance and $d$ tell us exactly where the first arm must point.  Since the central joint is a ball joint, we can choose the direction here by choosing any unit vector.  Thus, we have that $C(\Ess_d)$ is homeomorphic to $I \times S^2$ with $I$ being a closed interval.

\begin{figure}
\centering
\includegraphics[width = 0.3\textwidth]{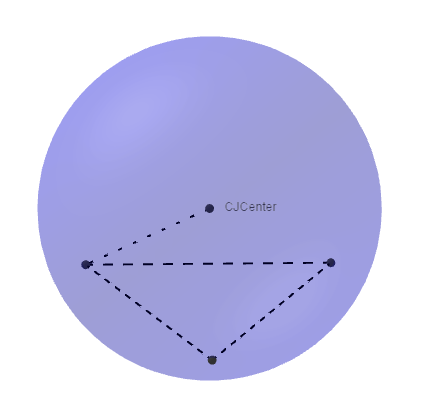}
\caption{\label{Workd}A slice of $W_{\Ess_d}$.   }
\end{figure}

One can readily describe the workspace of $\Ess_d$ as a union of spheres varying in radius and center.  One such sphere is depicted in Figure \ref{Workd}.  By inspection, we can see that the spheres appear to intersect, so we will determine where that intersection lies.  To assist our computations, we can first translate the joint so that the center of the base plate is at the origin and rotate so that the base plate sits on the $(x,y)$-plane.  Upon doing this, we can write the equation of each sphere as $x^2+y^2+(z-p)^2=d^2=p^2+b^2/3$ for a fixed plunge distance $p$.  We can rewrite this as $x^2+y^2+z^2-2pz=b^2/3$, then, we can solve for the intersection of two spheres with different plunge distances.  With some simple cancellation, we retrieve that either the plunge distances are equal or $z=0$.  This means that all of the spheres intersect at a common circle of radius $b/\sqrt{3}$ with center at the center of the base plate, and nowhere else.  This makes intuitive sense as we can always fold the arms so that the distal joint is located at the origin.  The circle comes from maintaining the angle with the plunge distance axis and rotating the ball joint.  Given this fact, we can uniquely determine the configuration for any point on the workspace that does not land on this circle.  Moreover, we can say that the inverse image of the circle is homeomorphic to $I \times S^1$, as for each point on the interval, we have a copy of the circle.
%-DMD-Explain why spheres intersect fully since we found it was surprising.
%--Analyze Factored Config Spaces \\

\subsection{Intersecting the Workspaces}

First, notice that each of the configuration spaces, $C(\Ess_\ell)$ and $C(\Ess_d)$, is determined by a single parameter and then a point on a sphere.  The parameter for $C(\Ess_\ell)$ is the base angle $\theta$, and the parameter for $C(\Ess_d)$ can be presented as either $d$ or $p$.  So, if we fix these two variables, we are left with two spheres in the configuration space which then intersect in the workspace.  Now, two spheres embedded in $\R^3$ can intersect in one of three ways: in a point, in a circle, or they are the same sphere.

By construction, both of the spheres intersect at the base hinge.  Thus, if the intersection of the two spheres is a point, it must be the base hinge.  This means that the only valid configuration for this is where all four of the bars are collinear and the length $\ell$ bars intersect and the length $d$ bars intersect.  This is a valid configuration, but it is a bit strange.  What this does yield, however, is a bound on the base angle $\theta$:

\begin{equation}
\theta \leq \pi + \arcsin\bigg(\frac{p}{d}\bigg).
\end{equation}

This means that we need only worry about this type of intersection when we have equality in the expression above.

We also note that the two spheres only have the opportunity to be identical in the case $d=\ell$.  However, setting $d=\ell$ is not a sufficient condition.  We also need the centers of the spheres to line up.  In other words, we need the vector from $S^1$ in $C(\Ess_\ell)$ to line up with the vector determined by $I$ in $C(\Ess_d)$.  This occurs when $\theta = \arcsin\big(\frac{p}{d}\big)$.  This might appear to be a minor case, but it can cause problems in our considerations for the configuration space of the Canfield Joint.

\begin{figure}
\centering
\includegraphics[width = 0.3\textwidth]{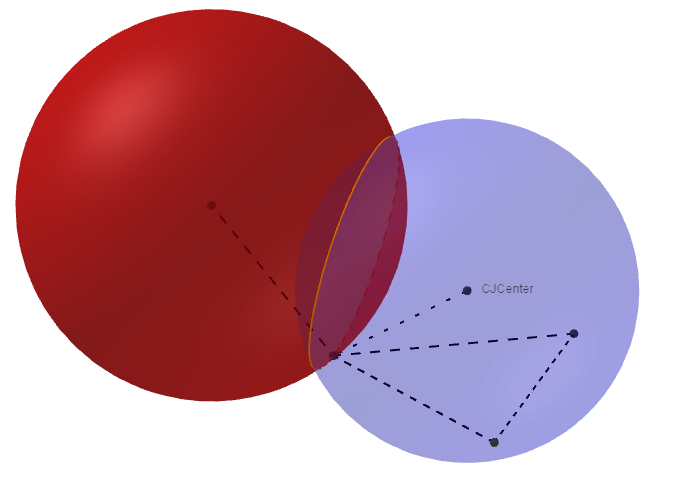}
\includegraphics[width = 0.3\textwidth]{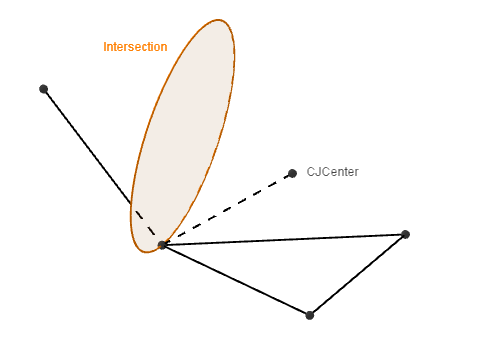}
\caption{\label{Intersect}When $W_{\Ess_\ell}$ and $W_{\Ess_d}$ intersect in a circle.   }

\end{figure}

\begin{figure}
\centering
\includegraphics[width = 0.4\textwidth]{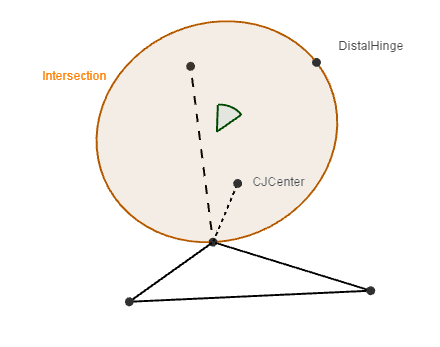}
\caption{\label{Phi}A visual representation of when $\phi \approx \frac{\pi}{4}$.   }
\end{figure}

When the $d$-sphere and the $\ell$-sphere intersect in a circle, we only need one more parameter to fix a location of the single arm workspace.  We can call this parameter $\phi$ as we only need an angle to locate ourselves on a circle.  We locate $\phi=0$ as the point on the circle furthest from the base hinge on the circle, and we define $\phi = \pm\pi$ when the distal hinge and the base hinge overlap, an illegal configuration.  Thus, we have that $-\pi < \phi < \pi$.  By imposing an orientation on the circle, we can then determine how $\phi$ moves the distal hinge.  As an example, Figure \ref{Phi} depicts when $\phi \approx \pi/4$.  With this parameter now defined, we can control the configuration of the 4-bar linkage using these three parameters, $\theta$, $p$, and $\phi$.
%-DMD-Describe construction of phi more precisely.

%--Show Intersection is what I think it is.\\
%--Justify Alternate Control Scheme \\

\subsection{Alternate Control Scheme}
%-DMD-This is the answer!  Make it clear!

Recall that the Canfield Joint itself required three parameters to control it, typically given as the three base angles $\theta_i$ for $i=1,2,3$.  Since this sublinkage requires three parameters to control it, one might think those three parameters could be construed into an alternative control scheme for the Canfield Joint in total.  It turns out that it can!

The control scheme determined by $(\theta, p, \phi)$ is particularly useful in answering our question about the kinematic workspace of a broken Canfield Joint.  Since our definition of an arm being broken is equivalent to saying that its base angle is fixed, fixing $\theta$ and varying the remaining two parameters, $p$ and $\phi$ yields a way to examine the boundaries of this workspace.  In section 5, we will explore some of the workspaces that are generated using this method.

\begin{figure}
\centering
\includegraphics[width = 0.4\textwidth]{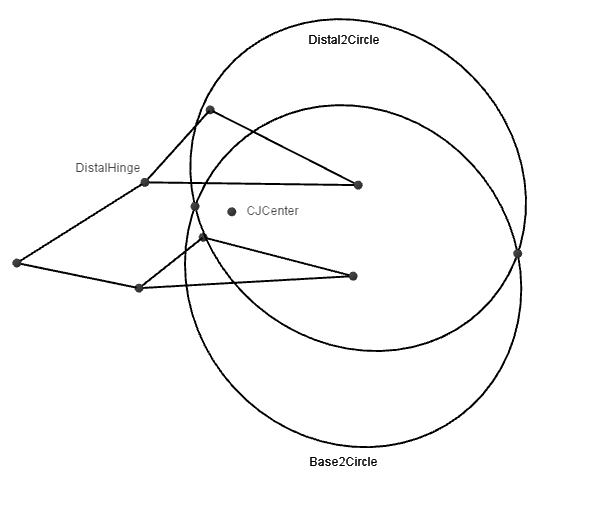}
\caption{\label{4Choice}A depiction of the possible locations for one of the remaining ball joints.   }
\end{figure}

We next need to construct the remainder of the Canfield Joint given the parameters $(\theta, p, \phi)$.  Since the remaining points on the distal plate and base plate are hinges, we can represent possible locations for the remaining ball joints via circles of radius $\ell$ centered at each hinge.  If the circles intersect, we have two possiblities.  In the case where there are two pairs of intersecting circles, there are at most four possible configurations of the Canfield Joint, provided the three parameters are known.  Such a choice for one of the arms is depicted in Figure \ref{4Choice}.  It is also possible for the circles to intersect in a point, which reduces the number of valid configurations by a factor of two for each arm that falls like this.  Finally, it is possible that the circles don't intersect.  This option is explored more in section 5.1.

The efficacy of this control scheme depends entirely on the bounds on the three parameters, $(\theta_1, p, \phi)$.  The bounds, however, will depend upon how the Canfield Joint is constructed.  The choice of parameters $b$ and $\ell$ are not only relevant to the bounds on these control parameters, but can also impact the topology of the configuration space of the Canfield Joint.  Some evidence for this is presented in the following section.

%--Construction of 4 possible CJ configs given control scheme. \\
%--Explain why choice matters \\

\subsection{A Preliminary Configuration Space Analysis}

Using the topological methodology outlined in section 3, we can loosely describe a decomposition of $C(\Ell)$, the configuration space of the four-bar linkage described section 4.1.  Future work may focus on fleshing out the connections in this configuration space analysis and describing the configuration space in more detail.

We begin by investigating what occurs when we fix two of our parameters, $\theta$ and $p$.  We will consider fixed values for $p$ and determine what happens as we vary $\theta$ within that space.  Upon doing this, we notice that there are three principal regimes with similar topological features.  These can be characterized as occurring when $d< \ell$, $d>\ell$, and $d=\ell$.

If we fix $d<\ell$ and vary $\theta$, we notice that the intersection is a circle that lies on the interior of $W_{\Ess_\ell}$.  As such, for every point except the origin, there are two configurations in $C(\Ess_\ell)$ and exactly one configuration in $C(\Ess_d)$ that correspond to that point in the workspace.  As such, there are two configurations in $C(\Ell)$ corresponding to every non-origin point in the workspace so long as $d<\ell$.

If we fix $d>\ell$ and vary $\theta$, we notice that the majority of the cases are similar to the previous case.  However, since $d>\ell$, we achieve a new case.  It is possible here for the intersection to correspond to a circle on the boundary of $W_{\Ess_\ell}$.  This occurs when $\theta = \arccos (b/(d\sqrt{3}))+\arccos (\ell/d)$, and there is a unique configuration in $C(\Ess_\ell)$ corresponding to the non-origin points of the circle.  As such, there is a unique configuration in $C(\Ell)$ corresponding to the non-origin points on that circle.  This feature is important as it gives a way to unify the two components we have considered so far away from the origin.  We will see this unification when we consider how to piece $C(\Ell)$ together.

Finally, if we fix $d=\ell$ and vary $\theta$, we again have that a majority of the cases resemble what occurs when $d<\ell$.  However, when $d=\ell$, there is again a case where the spheres intersect at the boundary of $W_{\Ess_\ell}$.  Unlike when $d>\ell$, though, we have that the intersection lies at the boundary only when $\theta = \arcsin\big(\frac{p}{d}\big)$.  But this is precisely when the spheres are identical!  As such, this particular configuration opens up a sphere's worth of possibilities.  The circle that lies on the boundary has a unique configuration in $C(\Ess_\ell)$ corresponding to it, but the remaining non-origin points on the sphere have two configurations in $C(\Ess_\ell)$ mapping to them just like the majority of the circles do.  This gives a very different structure to the regime $d=\ell$ that we will need to consider carefully when piecing $C(\Ell)$ together.

Before we attempt to piece $C(\Ell)$ together, there are two cases we have deliberately avoided.  The first is that we have avoided talking about the origin entirely.  Even though the origin is contained in all of the cases above, its inverse image is more subtle because of that.  On the one hand, in $C(\Ess_d)$, there is an interval of configurations that correspond to the origin (one for each value of $d$.  On the other hand, in $C(\Ess_\ell)$, there is a circle's worth of configurations corresponding to the origin as that is where all of the workspaces intersect.  This means that, in $C(\Ell)$, there are $I \times S^1$ configurations corresponding to the origin in the workspace.

The final case to consider is when $p=0$.  This case is subsumed by considering the origin since the only valid configuration here occurs when the plates have aligned, meaning that the distal hinge must land at the origin.  What matters here is in what regime this occurs.  When $p=0$, $d=b/\sqrt{3}$ is at its minimum value, meaning that the relationship between this value of $d$ and $\ell$ can affect the topology of the configuration space dramatically.  In particular, if $b/\sqrt{3} > \ell$, then we always have that $d>\ell$.  In this case, we can ignore the curious sphere that occurs in the workspace when $d=\ell$.  Notice that this relationship depends entirely on the measurements in the physical construction of the Canfield Joint ($b$ and $\ell$).  We then expect that there are three different possibilities:  $b/\sqrt{3}<\ell$, $b/\sqrt{3}=\ell$, and $b/\sqrt{3}>\ell$.

It is ultimately unsurprising that the topology of $C(\Ell)$ changes with respect to the measurements $b$ and $\ell$.  For other linkages, parameters measuring lengths of sides often determine topological differences in the configuration space.  We hope to improve our description of $C(\Ell)$ in future work.

\section{Results}

\begin{figure}
\centering
\includegraphics[width = 0.3\textwidth]{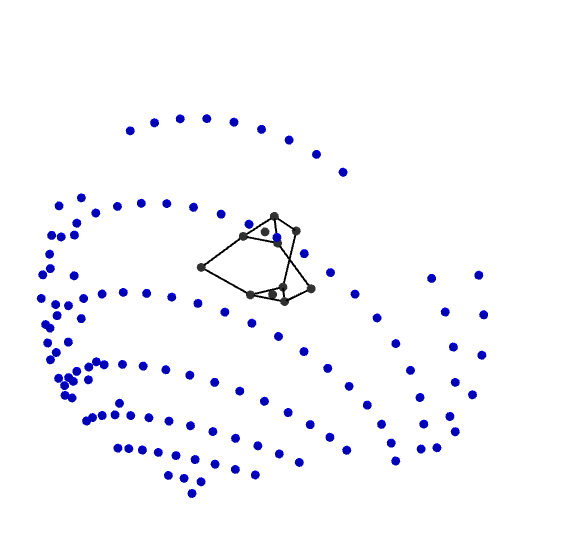}
\includegraphics[width = 0.3\textwidth]{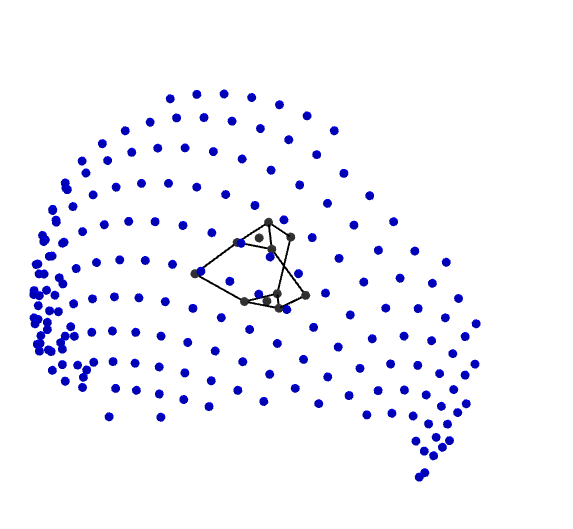}
\includegraphics[width = 0.3\textwidth]{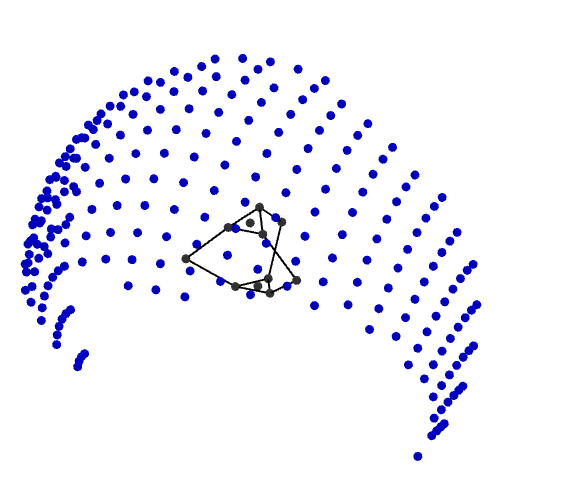} \\
\includegraphics[width = 0.3\textwidth]{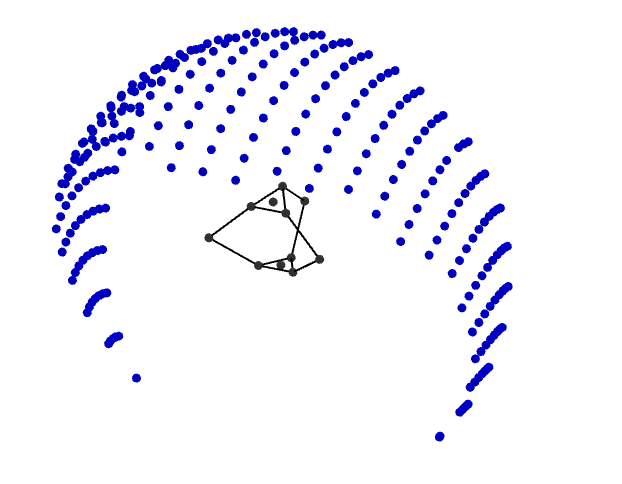}
\includegraphics[width = 0.3\textwidth]{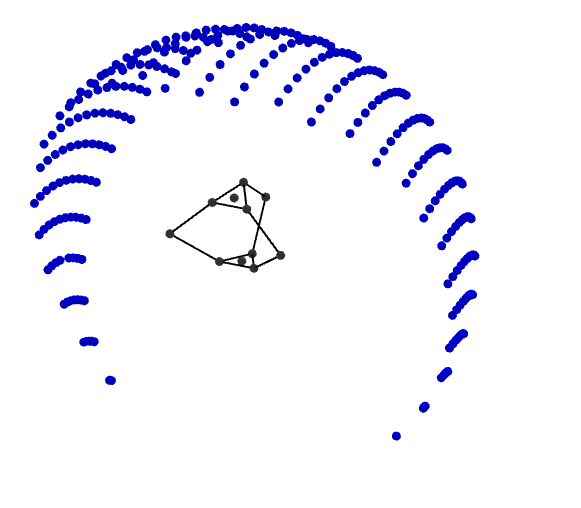}
\caption{\label{BWksp}Visualizations of the workspace of the center of the distal plate in the event that $\theta$ is fixed.  From left to right, we have:   Top: $\theta = 80^\circ,100^\circ,120^\circ$ \qquad Bottom: $\theta = 140^\circ,$ and $160^\circ$.    }
\end{figure}

After developing the necessary structure in GeoGebra, we used the software to generate images of some of workspaces of the center of the distal plate for the Canfield Joint when the base angle was fixed at various values.  The results of this are pictured in Figure \ref{BWksp}.

One of the first things that one might notice about these workspaces is that they all have a similar shape.  Each slice is reminiscent of the peel of an orange slice.  However, some of the angles have wider ranges than others.  For example, when the arm is stuck at $100^\circ$, we have nearly a quarter of a sphere's worth of motion.  In contrast, when the arm is stuck at $160^\circ$, we barely have a sixth of a sphere's worth of motion.  Another difference is the location of the workspace on the sphere, which varies with the angle in which the arm is locked.

It is worth noting that each of these pictures was generated with lots of hand calculations due to the lack of known physical bounds on the parameters.  Determining the physical bounds would allow us to construct more accurate visualizations, and it would allow us to control the Canfield Joint more effectively if such a failure mode is realized in practice.

\subsection{The Disconnect between $\Ell$ and the Canfield Joint}

\begin{figure}
\centering
\includegraphics[width = 1\textwidth]{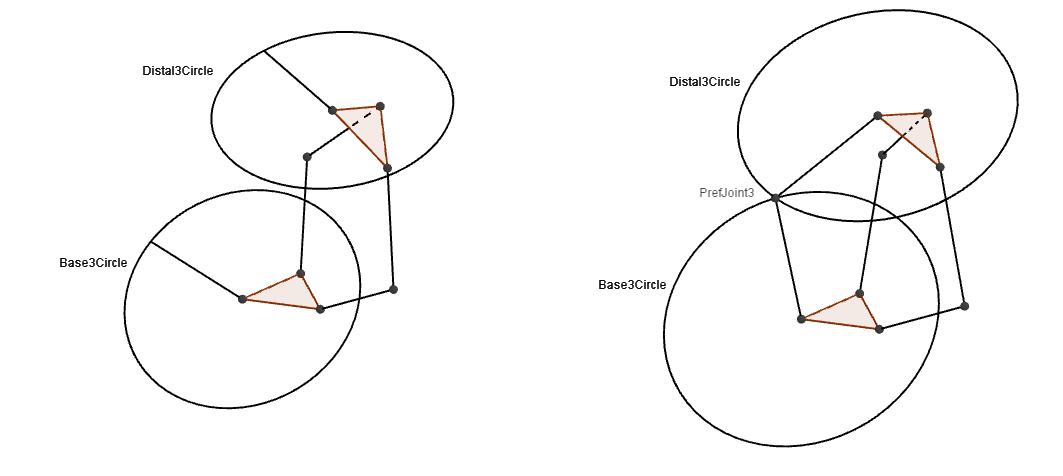}
\caption{\label{Invalid}The position $(150^\circ, 5.8, 20^\circ)$ in $C(\Ell)$ under two different $\ell$ and $b$ values.}
\end{figure}

Our final point in this section is to draw attention to an issue with attempting to control the Canfield Joint using the control scheme for $C(\Ell)$, in particular the $(\theta, p, \phi)$ control scheme.  When constructing the above results in GeoGebra, we witnessed an odd phenomena displayed in Figure \ref{Invalid}.  If we controlled the arm into a valid $C(\Ell)$ position when the plunge distance was nearly maximized, there were occasions when it would be impossible to construct the Canfield Joint.  We noticed that the relationship between $b$ and $\ell$ seemed to play a critical role in whether these configurations were valid or invalid.

As an example, in Figure \ref{Invalid} we construct the configuration $(150^\circ, 5.8, 20^\circ)$ in $C(\Ell)$ under two different measurements of $b$ and $\ell$.  On the left, $b=4$ and $\ell = 6$, and we see that the circles representing possible locations of the midpoint ball joint do not intersect.  On the right, $b=4$ and $\ell = 7$, and we have two points of intersection of the circles, as we expect to occur.  Part of why this happens is that 5.8 is much closer to the maximum in the case on the left as compared to the case on the right.  Even with this explanation, that we have the disconnect on the left at all displays the naivety of our current bounds on the control parameters of $C(\Ell)$.  Extending these to the Canfield Joint may involve much more subtle bounds than what we have come up with, and that is worth exploring further.

\section{Conclusion}
While the results of this paper were motivated by the iROC project, it is worthwhile to note that these results can apply to any application in which the Canfield Joint is used.  We note that NASA's first interaction with the Canfield Joint was in using it to point propulsion systems.  Using a propulsion system may well interfere with the ``broken'' arm in the model.  Specifying how a propulsion system might change the arm would likely depend on the specific construction of the Canfield Joint involved.

%--Applications to anything w/ CJ \\

The results of this paper can be expanded by finding explicit equations relating the $(\theta,p,\phi)$ control scheme to the $(\theta_1,\theta_2,\theta_3)$ control scheme.  Since this work was primarily done in GeoGebra, the relationships are apparent, but not explicit.  It would be valuable in terms of both computing physical bounds on the parameters as well as giving programmers a way to use this alternate control scheme to effectively move the Canfield Joint.  One idea that is mentioned in section 3.4 of \cite{Canfield} is moving the joint while holding $p$ fixed.  This is quite challenging to do with the $(\theta_1, \theta_2, \theta_3)$ control scheme, but should be trivial in $(\theta, p, \phi)$ once the specific forward kinematics are known.
%--Finding Explicit Eq'ns \\
%--Determining Bounds on Parameters \\ 

The field of topological robotics was recently named in Michael Farber's book \cite{Farber}, and it is a relatively new addition to the applied topology arsenal.  As a final note, we believe the techniques of topological robotics could be used in plenty of different applications.  We hope that opening this door a crack may lead to greater collaboration between topologists and engineers which may be beneficial to both disciplines.
%--Performing Top'l Analysis on other Linkages


\begin{thebibliography}{1}

\bibitem[Can]{Canfield}
Stephen L Canfield. {\em Development of the Carpal Wrist; a Symmetric, Parallel-Architecture Robotic Wrist}. PhD thesis, Virginia Polytechnic Institute and State University. 1997

\bibitem[Wal]{Walker} Kevin Walker. {\em Configuration Spaces of Linkages}. Undergraduate Thesis, Princeton University.  1985

\bibitem[Far]{Farber} Michael Farber. {\em Invitation to Topological Robotics}. European Mathematical Society. 2008

\end{thebibliography}
\end{document}